\documentclass[sigconf]{acmart}

\AtBeginDocument{%
  \providecommand\BibTeX{{%
    \normalfont B\kern-0.5em{\scshape i\kern-0.25em b}\kern-0.8em\TeX}}}

\usepackage{url}  

\graphicspath{{images/}}
\usepackage{amsmath}
\usepackage{subfig}

\usepackage{multirow}
\usepackage{algorithm}
\usepackage{bm}
\usepackage{soul}
\usepackage{color}
\definecolor{Gray}{gray}{0.935}
\usepackage{booktabs}
\usepackage{hyperref}
\usepackage{soul}

\definecolor{WIP}{rgb}{0.875,0.398,0.398}
\definecolor{SIP}{RGB}{255,0,0}
\definecolor{WER}{rgb}{0.426,0.617,0.918}
\definecolor{SER}{rgb}{0.066,0.332,0.797}
\definecolor{WEX}{rgb}{0.574,0.765,0.488}
\definecolor{SEX}{rgb}{0.218,0.461,0.113}

\definecolor{light_blue}{RGB}{0,85,149}

\definecolor{specific}{RGB}{106,168,79}

\newcommand{\xhdr}[1]{\vspace{1.0mm}\noindent{{\bf #1.}}}

\newcommand{\ourmodel}{\textsc{Partner}~}

\newcommand\textb[1]{{#1}}

\setcopyright{iw3c2w3}

\begin{document}

\title[Towards Facilitating Empathic Conversations in Online Mental Health Support]{Towards Facilitating Empathic Conversations in Online Mental Health Support: A Reinforcement Learning Approach}






\author{Ashish Sharma$^{\spadesuit}$ \: \: \: Inna W. Lin$^{\spadesuit}$ \: \: \: Adam S. Miner$^{\clubsuit\heartsuit}$ \: \: \: David C. Atkins$^\diamondsuit$ \: \: \: Tim Althoff$^{\spadesuit}$}
\affiliation{
  \institution{$^\spadesuit$Paul G. Allen School of Computer Science \& Engineering, University of Washington \\
  $^\clubsuit$Department of Psychiatry and Behavioral Sciences, Stanford University \\
  $^\heartsuit$Center for Biomedical Informatics Research, Stanford University  \\
  $^\diamondsuit$Department of Psychiatry and Behavioral Sciences, University of Washington}
  \city{}
  \state{}
  \country{}
}
\email{{ashshar, ilin, althoff}@cs.washington.edu}

\renewcommand{\authors}{Ashish Sharma, Inna W. Lin, Adam S. Miner, David C. Atkins, Tim Althoff}
\renewcommand{\shortauthors}{Sharma, et al.}

\begin{abstract}
  Online peer-to-peer support platforms enable conversations between millions of people who seek and provide mental health support. If successful, web-based mental health conversations could improve access to treatment and reduce the global disease burden. Psychologists have repeatedly demonstrated that \textit{empathy}, the ability to understand and feel the emotions and experiences of others, is a key component leading to positive outcomes in supportive conversations. However, recent studies have shown that highly empathic conversations are rare in online mental health platforms. 
  
  In this paper, we work towards improving empathy in online mental health support conversations. We introduce a new task of \textit{empathic rewriting} which aims to transform low-empathy conversational posts to higher empathy. Learning such transformations is challenging and requires a deep understanding of empathy while maintaining conversation quality through text fluency and specificity to the conversational context.
  Here we propose \textsc{Partner}, a deep reinforcement learning (RL) agent that learns to make sentence-level edits to posts in order to increase the expressed level of empathy while maintaining conversation quality. Our RL agent leverages a policy network, based on a transformer language model adapted from GPT-2, which performs the dual task of generating candidate empathic sentences and adding those sentences at appropriate positions. 
  During training, we reward transformations that increase empathy in posts while maintaining text fluency, context specificity, and diversity. Through a combination of automatic and human evaluation, we demonstrate that \ourmodel successfully generates more empathic, specific, and diverse responses and outperforms NLP methods from related tasks such as style transfer and empathic dialogue generation. This work has direct implications for facilitating empathic conversations on web-based platforms. 
\end{abstract}




\copyrightyear{2021}
\acmYear{2021}
\acmConference[WWW '21]{Proceedings of the Web Conference 2021}{April 19--23, 2021}{Ljubljana, Slovenia}
\acmBooktitle{Proceedings of the Web Conference 2021 (WWW '21), April 19--23, 2021,
Ljubljana, Slovenia}
\acmPrice{}
\acmDOI{10.1145/3442381.3450097}
\acmISBN{978-1-4503-8312-7/21/04}

\settopmatter{printacmref=true}

\maketitle

\section{Introduction}

\begin{figure}[t]
\centering
\includegraphics[width=\columnwidth]{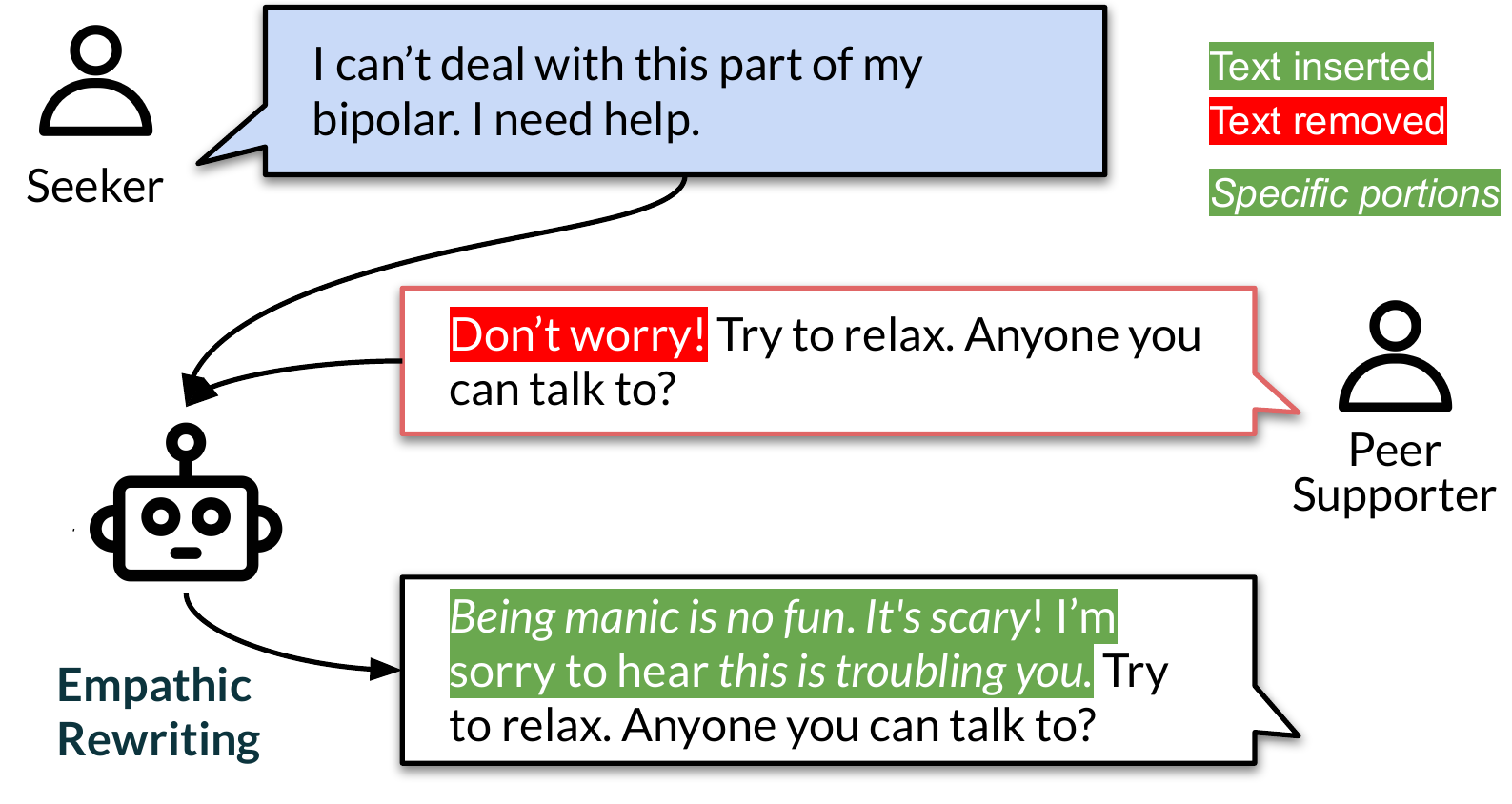}
\vspace{-20pt}
\caption{An overview of the empathic rewriting task. Given a post from support seeker and a low-empathy response, the task is to rewrite the response for making it more empathic, through text \textcolor{specific}{insertions} and \textcolor{red}{deletions}. This task requires inferring \textcolor{specific}{\textit{specific}} feelings and experiences from seeker's post and using them for making appropriate changes to the response through empathic mechanisms like emotional reactions, interpretations, and explorations~\cite{sharma2020computational}. \textit{Examples in this paper have been paraphrased for anonymization~\cite{matthews2017stories}.}}
\label{fig:example}
\vspace{-15pt}
\end{figure}

Online mental health support platforms such as TalkLife (\href{https://talklife.co}{\textcolor{light_blue}{talklife.co}}) are used by millions of users for expressing emotions, sharing stigmatized experiences, and receiving peer support. 
These platforms might help improve access to mental health support as mental health care remains a global challenge with widespread shortages of workforce~\cite{olfson2016building}, limited in-person treatment options, and other barriers like stigma~\cite{white2001receiving}.
A key component of providing successful support is \textit{empathy}, the ability to understand or feel the emotions and experiences of others~\cite{elliott2011empathy}. Quantitative evidence shows that empathic interactions have strong associations with symptom improvement in mental health support~\cite{elliott2018therapist} and are instrumental in building therapeutic alliance and rapport~\cite{bohart2002empathy,robert2011empathy}. Yet, highly empathic conversations are rare on online support platforms~\cite{sharma2020computational}. 

Empowering peer supporters on online support platforms with feedback and training, for example through machine-in-the-loop writing systems~\cite{clark2018creative,tanana2019development}, has the potential to help supporters express higher levels of empathy and in turn improve the effectiveness of these platforms~\cite{miner2019key,imel2015computational,sharma2020computational}. Traditional methods for training empathy (e.g., in-person counselor training) do not scale to the millions of users of online support platforms. However, computational methods that can support peer-supporters by suggesting ways to modify existing conversation utterances to make them more empathic may help meet this need of feedback and training and indirectly benefit support seekers on the platform.

In this paper, we introduce \textbf{Empathic Rewriting}, a new task that aims to transform low-empathy conversations to higher empathy (Figure~\ref{fig:example}). For example, given a post from a support seeker "\textit{I can’t deal with this part of my bipolar. I need help.}" and a low-empathy response "\textit{Don't worry! Try to relax. Anyone you can talk to?}", we want to increase empathy in the response by transforming it to "\textit{Being Manic is no fun. It's scary! I'm sorry to hear this is troubling you. Try to relax. Anyone you can talk to?}"; the rewritten response should communicate more empathy through an understanding of feelings and experiences ("\textit{Being manic is no fun. It's scary}") and display of felt emotions ("\textit{I'm sorry to hear this is troubling you}").

Performing such transformations is a challenging task: First, empathy is a complex, conceptually nuanced construct and requires understanding the feelings and experiences shared by the support seeker. In the example above, one needs to understand that being "\textit{bipolar}" can be "\textit{scary}", involves "\textit{manic}" phases, and communicate this in the response. Second, for empathic rewriting to be purposeful, it should not undermine other conversation goals like language fluency, context specificity, and diversity. Making changes that lead to ungrammatical posts with empathic portions (e.g., "\textit{Scary it is manic being}") may not be helpful and obstruct useful feedback. Further, making the same transformation to every response (e.g., rewrite every response to "\textit{I understand how you feel}") would lead to non-specific and generic responses reducing the overall conversational quality~\cite{see2019makes,li2016diversity}. Third, the task of empathic rewriting requires changes that go beyond simple word-level transformations, often requiring multiple new sentences to be added or replaced (e.g., three sentence insertions and one sentence removal in the example in Figure~\ref{fig:example}). This is different from related style transfer tasks~\cite{shen2017style,li2018delete} where even changing a single word may suffice for transferring from negative to positive sentiment (e.g., replace "\textit{bad}" with "\textit{good}" in the sentence "\textit{the movie was bad}"). Finally, supervised methods commonly used for similar tasks such as style transfer~\cite{shen2017style,li2018delete} and content debiasing~\cite{pryzant2020automatically,ma2020powertransformer} usually require a large parallel dataset. Such a dataset is not yet available for empathic rewriting and hard to collect as it would require a large number of clinical psychologists and counselors well-versed in the complex construct of empathy.

To address the challenges described above, we propose \textsc{Partner},\footnote{em\textbf{PA}thic \textbf{R}ewri\textbf{T}ing in me\textbf{N}tal h\textbf{E}alth suppo\textbf{R}t} a deep reinforcement learning (RL) model for the task of empathic rewriting (Section~\ref{sec:model}). We design an RL agent which learns to add new empathic sentences to posts or replace existing sentences in posts with more empathic ones. The agent operates on a pair of seeker post and the original response post (which rarely is highly empathic~\cite{sharma2020computational}) and makes edits to the response at the level of a sentence by simultaneously (a) identifying positions in the original response post where changes are required, and (b) generating empathic sentences for insertion or replacement at the identified positions (Section~\ref{subsec:actions}). We model this agent using a policy network based on a transformer decoder model adapted from GPT-2~\cite{radford2019language}. We build upon existing large-scale pre-training of GPT-2 on conversations, as done in DialoGPT~\cite{zhang2019dialogpt}, and modify it to perform the two simultaneous actions of identifying positions and generating empathic sentences for empathic rewriting (Section~\ref{subsec:policy}). Through carefully constructed scoring functions, we reward transformations that increase empathy in posts while maintaining text fluency, context specificity, and diversity (Section~\ref{subsec:rewards}). 

Evaluating complex conversational constructs such as empathy is fundamentally challenging~\cite{sharma2020computational}. Therefore, we combine comprehensive automatic evaluation with expert-based human evaluation. Our experiments demonstrate that \ourmodel can effectively increase empathy in posts in fluent, specific, and diverse ways and outperforms baselines used in related text generation tasks by $>35\%$ in empathy improvement (Section~\ref{sec:experiments}). Also, 
\ourmodel is the only approach that consistently improves empathy and does not lead to a \emph{loss} of empathy when rewriting an already highly empathic post, while all baselines tend to propose a large number of edits that only make the situation worse (Section~\ref{subsec:automatic-metrics}).
Lastly, through comprehensive human evaluation, we show that experts in clinical psychology prefer rewritings of \ourmodel compared to baselines, based on empathy, specificity, and fluency (Section~\ref{subsec:human-eval}). 
We view our approach and findings as a key step towards building AI systems for facilitating empathic conversations on online mental health support platforms, but these insights may generalize beyond mental health to other conversational settings on web-based platforms. We share our code publicly at \href{https://github.com/behavioral-data/PARTNER}{\textcolor{light_blue}{https://github.com/behavioral-data/PARTNER}}.

\section{Related Work}

We build upon prior work on NLP for online mental health support, empathic dialogue generation, reinforcement learning for text rewriting and natural language generation, and AI-assisted writing.

\subsection{NLP for online mental health support}
Broadly, our work relates to existing research on NLP for online mental health support. These efforts have predominantly focused on analyzing techniques that are effective for seeking and providing conversational support such as adaptability to various contexts and diversity of responses~\cite{althoff2016large,perez2019makes,zhang2020balancing,sharma2018mental,yang2019channel}. \textb{Researchers have also built methods for identifying therapeutic actions~\cite{lee2019identifying}, quantifying language development of counselors~\cite{zhang2019finding}, extracting patterns of conversational engagement~\cite{sharma2020engagement}, analyzing moderation~\cite{wadden2020effect}, and detecting cognitive restructuring~\cite{pruksachatkun2019moments} in supportive conversations.} Here, we focus on a particular conversation technique, \textit{empathy}, which is key in counseling and mental health support~\cite{castonguay2017and,elliott2011empathy}. Our work builds on previous efforts on understanding and building computational methods for identifying empathy in online health communities~\cite{khanpour2017identifying}, face-to-face therapy~\cite{gibson2016deep,perez2017understanding}, and text-based peer-to-peer support~\cite{sharma2020computational}. We extend this work by learning to improve empathy in online mental health support conversations through a reinforcement learning method for empathic rewriting (Section~\ref{sec:model}).

\subsection{Empathic dialogue generation}
Our task of empathic rewriting is related to empathic dialogue generation but has a key difference as it involves making empathic changes to \emph{existing} responses instead of generating new responses from scratch. While research on generating empathic dialogue has mainly focused on chit-chat, open-domain conversations~\cite{rashkin2019towards,lin2019moel,majumder2020mime}, we work on conversations in online mental health support. Moreover, most empathic dialogue generation methods have a tendency of enabling empathic conversations through emotional grounding~\cite{rashkin2019towards} or emotion mimicking~\cite{majumder2020mime}. In mental health support, however, communicating the cognitive aspects of empathy, related to understanding the experiences and feelings of others, are more valued by mental health professionals~\cite{sharma2020computational,truax1967modern,selman1980growth}. We extend this work with the task of empathic rewriting (Section~\ref{sec:problem}) and by leveraging both emotional and cognitive aspects of empathy, using a theoretically-grounded framework of empathy~\cite{sharma2020computational} (Section~\ref{sec:model}).

\subsection{Text rewriting and AI-assisted systems}
Text rewriting is a broad subarea in natural language processing that includes tasks such as style transfer~\cite{shen2017style,li2018delete}, content debiasing~\cite{pryzant2020automatically,ma2020powertransformer}, and controllable text generation~\cite{hu2017toward,dathathri2019plug,mai2020plug}. We propose empathic rewriting as a new text rewriting task in which conversational utterances are rewritten for increasing them in empathy (Section~\ref{sec:problem}). This task presents unique challenges different from other text rewriting tasks: it requires understanding empathy in conversational contexts and leveraging that understanding for making empathic changes while ensuring high conversational quality in terms of language fluency, context specificity, and diversity. 

Here, we propose a reinforcement learning (RL) model for the task of empathic rewriting (Section~\ref{sec:model}). Previous work has used RL for the task of sentiment transfer~\cite{luo2019dual} by only using text generations as actions. Here, we design an RL agent that simultaneously learns to (a) identify positions for making improvements and (b) generating empathic sentences for insertion or replacement at the identified positions. These actions are important because the task of empathic rewriting requires changes that go beyond simple word-level transformations, as common in sentiment transfer tasks (e.g., change "\textit{bland}" to "\textit{delicious}" in "\textit{the food was bland}" for transferring from negative to positive sentiment).

Prior work has built systems that leverage identification of effective conversational strategies such as asking open-ended questions for training users in counseling~\cite{huang2020challenges}. Computational methods that can perform empathic rewriting can be used for suggesting ways to make conversations more empathic in similar feedback and training systems for mental health support and counseling. In related context, researchers have built AI tools for writing assistance in negotiations~\cite{zhou2019dynamic}, composing emails~\cite{chen2019gmail}, language translation~\cite{santy2019inmt}, creative writing~\cite{clark2018creative}, and communication of politeness~\cite{fupoliteness}.

\vspace{-5pt}
\section{Dataset Description}
\label{sec:dataset}
In this section, we describe the dataset used for the task of empathic rewriting.


\subsection{The TalkLife platform}
\label{subsec:talklife}
TalkLife (\href{https://talklife.co}{\textcolor{light_blue}{talklife.co}}) is the largest online peer-to-peer support platform for mental health support. It enables conversations between people seeking support (\textit{support seekers}) and people providing support (\textit{peer supporters}) in a thread-like setting. We call the post authored by a support seeker as \textit{seeker post}, and the response by a peer supporter as \textit{response post}. Table~\ref{tab:dataset} describes the statistics of conversational threads on the TalkLife platform.

\xhdr{Curating mental health-related conversations} As noted by Sharma et al.~\cite{sharma2020computational}, the TalkLife platform hosts a significant number of common social media interactions (e.g., \textit{Happy mother's day}). Here, we focus our analyses on mental health-related conversations and filter out such posts. We manually annotate $\sim$3k posts with answers to the question \textit{"Is the seeker talking about a mental health related issue or situation in his/her post?"}. Using this annotated dataset, we train a standard text classifier based on BERT~\cite{devlin2018bert} (achieving an accuracy of $\sim$85\%). We apply this classifier to the entire TalkLife dataset and create a filtered dataset of mental health-related conversations. This dataset contains 3.33M interactions from 1.48M seeker posts.

 \begin{table}[t]
 \centering
\begin{tabular}{p{0.45\columnwidth}p{0.35\columnwidth}}
\toprule

\textbf{Dataset Statistics} &  {TalkLife}                            \\
\midrule
\# of Seeker posts & 10.9M \\
\# of Response posts & 26.9M \\
\# of Users & 642K  \\
Observation Period & May 2012 to June 2020 \\
\bottomrule
\end{tabular}
 \caption{Statistics of the TalkLife dataset.}
   \vspace{-20pt}
  \label{tab:dataset}
 \end{table}

\begin{figure}[t!]
\centering
\vspace{-5pt}
\includegraphics[width=\columnwidth]{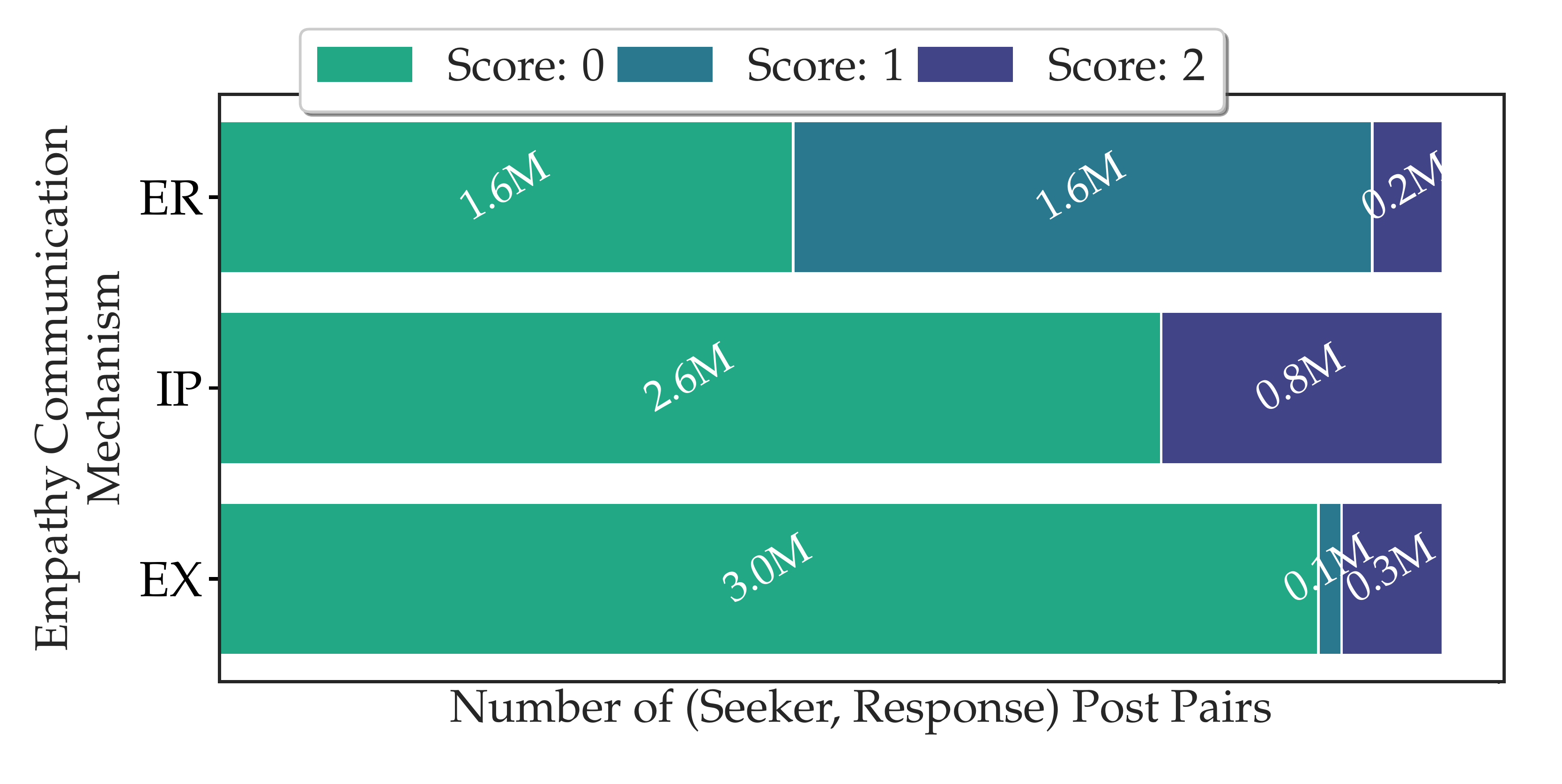}
\vspace{-25pt}
\caption{Expression of high levels of empathy is very low in online support platforms, especially for Interpretations (IP) and Explorations (EX). Emotional reactions (ER) are slightly more common.}
\label{fig:empathy-dataset}
\vspace{-15pt}
\end{figure}

\vspace{-5pt}

\subsection{Creating a dataset of empathic posts}
\label{subsec:empathic-data}
Training supervised methods would require a large parallel dataset of corresponding pairs of posts with low and high empathy, respectively. As empathy is a complex phenomenon, collecting such a dataset is challenging and would likely require psychology experts. Here, we create a large non-parallel dataset with empathy measurements for training unsupervised and self-supervised computational models and a small parallel dataset with expert empathic rewritings for conducting evaluations.

\xhdr{Computational labeling with empathy measurements} 
We computationally label our dataset of 3.33M interactions with empathy measurements using a recently proposed framework of expressed empathy in mental health support~\cite{sharma2020computational}. This framework consists of three empathy communication mechanisms -- (1) \textit{Emotional Reactions} (expressing emotions such as warmth, compassion), (2) \textit{Interpretations} (communicating an understanding of feelings and experiences), and (3) \textit{Explorations} (improving understanding of the seeker by exploring feelings and experiences). For each communication mechanism, the authors design a three-point scale (0 to 2). We computationally label all pairs of (seeker post, response post) in our dataset based on this empathy scale. For this, we use a classification model (RoBERTa-based, bi-encoder attention with an accuracy of $\sim$80\%) developed by Sharma et al.~\cite{sharma2020computational}. Figure~\ref{fig:empathy-dataset} shows the statistics which indicate that high levels of empathy expressions are uncommon in online support platforms, highlighting the need for building systems for improving empathy (e.g., through feedback using empathic rewriting (Section~\ref{sec:problem})).  We use this dataset for a supervised warm-start training in our reinforcement learning model (Section~\ref{subsec:optimization}) and for training unsupervised baselines (Section~\ref{subsec:baselines}).

\xhdr{Expert empathic rewritings}
Additionally, we create a small parallel dataset of 180 pairs of corresponding low and rewritten high empathy response posts with rewritings from people having substantial expertise in empathy, mental health, and therapy (six graduate students in clinical psychology; none are co-authors). We showed them pairs of seeker and response posts and asked them to modify the response post for improving it in empathy. This expert-based dataset is designed to represent the best possible responses and we use it as ground truth for evaluation (Section~\ref{subsec:human-eval}).

\vspace{-5pt}

\subsection{Privacy, ethics, and disclosure}
The dataset was sourced with license and consent from the TalkLife platform. All personally identifiable information
(user and platform identifiers) in our dataset was removed. This work was approved by University of Washington's Institutional Review Board. We do not make any treatment recommendations or diagnostic claims. 

\xhdr{Towards preventing unsafe rewritings} 
We acknowledge that building computational models for intervention in high-stakes settings such as mental health necessitates ethical considerations. There is a risk that in attempting to help, responses could have the opposite effect, which could be deadly in cases of self-harm. No current computational approach will identify and respond to harm-related utterances perfectly~\cite{miner2020assessing}. Thus, risk mitigation steps are appropriate in this context. Here, we remove all posts that contain a pre-defined unsafe regular expression (e.g., \textit{$*$commit suicide$*$}) from our analyses and training in collaboration with mental health professionals. Future work testing or deploying AI systems should assess safety-related risk, and also potential sources of bias (e.g., race, ethnicity, age, or gender bias in training data or models).

\section{Problem Definition and Goals}
\label{sec:problem}
In this section, we formulate the task of empathic rewriting and state the associated goals.

\vspace{-5pt}

\subsection{Empathic Rewriting}
\label{subsec:empathic-rewriting}
We introduce \textit{empathic rewriting}, a new task that aims to transform low-empathy conversational posts to higher empathy. In contrast with empathic dialogue generation~\cite{rashkin2019towards,lin2019moel,majumder2020mime}, where the objective is to generate empathic posts from scratch, this task requires making changes to existing posts in order to make them empathic. This is more consistent with realistic use-cases in difficult, high-stakes settings such as online support systems, which are likely to augment, rather than replace humans~\cite{miner2019key}.

Formally, let $\mathbf{S_{i}}$ be a seeker post and $\mathbf{R_{i}}$ be a corresponding response post. We aim to transform $\mathbf{R_{i}}$ into its more empathic counterpart $\mathbf{\hat{R}_{i}}$. 

\begin{figure*}[t]
\centering
\vspace{-5pt}
\includegraphics[width=0.99\linewidth]{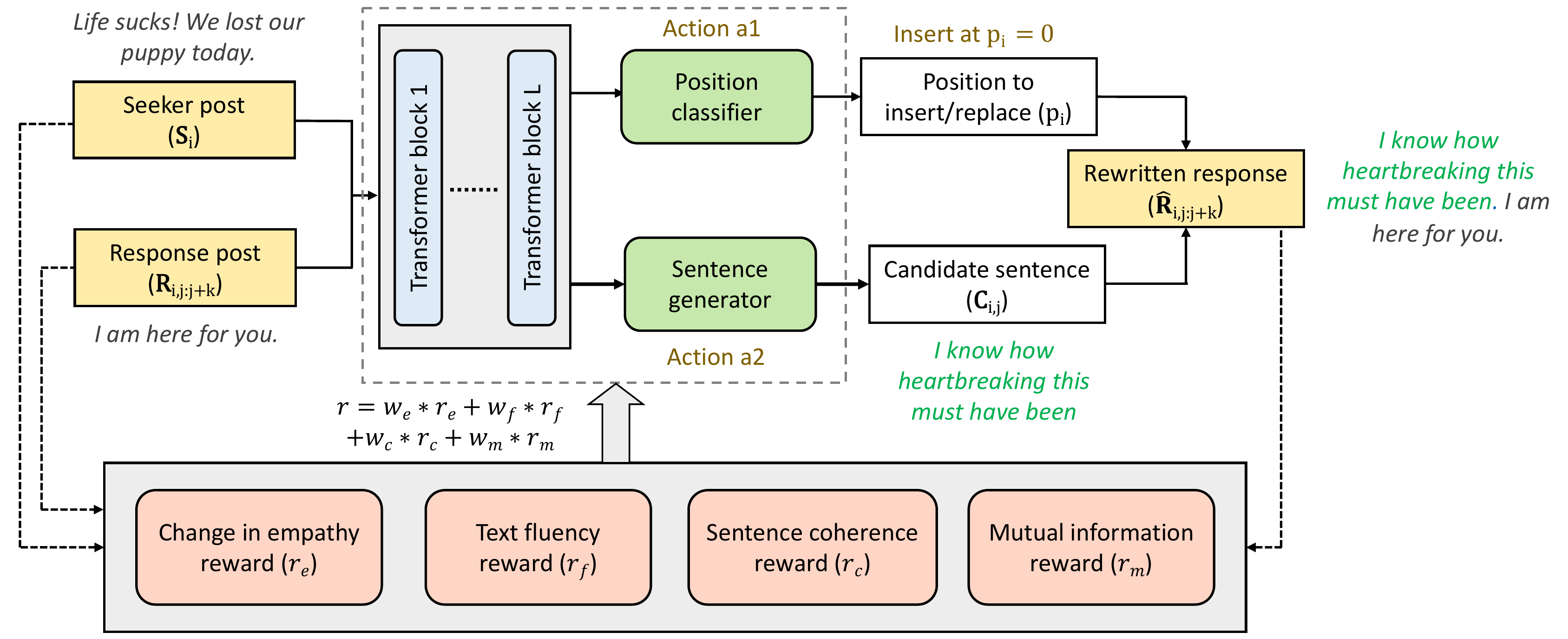}
\vspace{-10pt}
\caption{\ourmodel uses a deep reinforcement learning approach for Empathic Rewriting. It leverages a transformer language model for performing the two actions of (1) selecting positions for insertion or replacement and (2) generating candidate empathic sentences. It uses four reward functions that promote increase in empathy, text fluency, sentence coherence, context specificity, and diversity.}
\label{fig:model}
\vspace{-10pt}
\end{figure*}

\vspace{-5pt}
\subsection{Goals}
\label{subsec:goals}
For empathic rewriting to be useful in improving mental health support conversations, the rewriting process should achieve specific goals related to empathy, conversation and natural language generation quality, and purposeful and precise feedback:

\xhdr{Theoretically-grounded empathy} 
Empathy is complex and conceptually nuanced; over time psychology research has emphasized multiple aspects of empathy~\cite{bohart1997empathy,duan1996current,batson2009these,davis1980multidimensional}. For example, computational research typically defines empathy as reacting with emotions of warmth and compassion~\cite{buechel2018modeling}. However, psychotherapy research emphasizes aspects of empathy related to communicating cognitive understanding of feelings and experiences of others ~\cite{selman1980growth}. For empathic rewriting to be useful and potentially adopted in online mental health support, we need to design methods grounded in psychology and psychotherapy research. Here, we adopt the theoretically-grounded framework of empathy designed by Sharma et al.~\cite{sharma2020computational}. We leverage empathy measurements based on this framework as (1) reward signals in our model for empathic rewriting (Section~\ref{subsec:rewards}), and (2) an automatic evaluation metric for judging improvements in empathy from various rewriting models (Section~\ref{subsec:automatic-results}).

\xhdr{Context specificity and response diversity} 
Consider a rewriting approach that transforms every response to a generic but empathic response (e.g., "\textit{That must have been really hard for you}"). While this approach may seem to "solve" empathic rewriting, it suffers from two key issues. First, the responses generated by this approach would lack specificity to the emotions and experiences shared in the seeker post, which is important for empathy and effective mental health support~\cite{robert2011empathy,majumder2020mime}. Second, performing this same transformation to millions of responses on online platforms would dramatically reduce response diversity which has been shown to be important for mental health support~\cite{althoff2016large} as well as in general dialogue research~\cite{see2019makes,li2016diversity}. 

Thus, the task of empathic rewriting interplays with other issues related to conversation and natural language generation quality and effective mental health support. Ensuring that the rewritten response is specific and diverse, along with empathic is challenging but critical for obtaining purposeful transformations. In this work, we learn rewriting actions that simultaneously achieve the goals of context specificity and response diversity using a reinforcement learning approach (Section~\ref{subsec:rewards}) and we evaluate these goals using a combination of automatic and human evaluation (Section~\ref{subsec:automatic-results},\ref{subsec:human-eval}).

\xhdr{Text fluency and sentence coherence} 
In addition, only generating empathic words or phrases may not be sufficient. Without appropriate measures, the rewriting process may lead to an ungrammatical, non-fluent final response (e.g., "\textit{Scary being is it manic}"). Also, making changes that are incoherent with the original response may not be appropriate (e.g., changing "\textit{Sorry to hear that you lost your job. I hope you get a new job soon.}" to "\textit{Sorry to hear that you lost your job. Congrats on your job promotion. I hope you get a new job soon.}"). In this paper, we avoid such responses with non-fluent and incoherent portions through carefully constructed reward functions (Section~\ref{subsec:rewards}) and conduct both automatic and human evaluations of models on text fluency and sentence coherence (Section~\ref{subsec:automatic-results},\ref{subsec:human-eval}).

\xhdr{Rewriting for feedback and training} 
An important way in which the task of empathic rewriting can be used is for providing feedback and training to people through machine-in-the-loop writing systems~\cite{clark2018creative,tanana2019development}. For humans to adopt such feedback, however, the rewriting process should make changes that are precise and specific to the original response. This means that the number of changes should be kept minimal and that the changes themselves should be suitable to the original response. For example, adding 10 sentences to a one-sentence response may not be useful. Here, we train a reinforcement learning agent which learns when to stop making changes through a special "stopping" action (Section~\ref{subsec:actions}). We evaluate the number of transformations different models need for empathic rewriting through a standard edit-distance based scoring metric (Section~\ref{subsec:automatic-results}).

\section{\textsc{Partner}: Empathic rewriting using reinforcement learning}
\label{sec:model}

Here, we present \textsc{Partner}, a reinforcement learning model for the task of empathic rewriting. We first explain the general reinforcement learning framework and its applicability to our setting. We then describe the various components of our model (states, actions, policy, and rewards) and our training strategy.

\subsection{Reinforcement Learning Framework}
\label{subsec:rl_framework}
We adopt the standard reinforcement learning framework consisting of a collection of states $\mathcal{S}$, a set of actions $\mathcal{A}$, a policy $\mathcal{\pi}$, and rewards $\mathcal{R}$~\cite{sutton2018reinforcement}. In this framework, given a state $s \in \mathcal{S}$, an agent takes an action $a \in \mathcal{A}$ according to the policy $\mathcal{\pi} : \mathcal{S} \times \mathcal{A} \to [0,1]$. The policy defines whether the agent should take action $a$ in a state $s$. The goal of the reinforcement learning agent is to learn a policy which maximizes the reward $r:\mathcal{S} \times \mathcal{A} \to \mathcal{R}$. 

Here, we design a reinforcement learning model for the task of empathic rewriting. Conceptually, our agent leverages context from the seeker post which it uses for making specific empathic changes. Alongside, it operates on the response post, looks for areas where empathy could be improved, and works on those improvements in fluent, coherent, specific, and diverse ways. Moreover, it ensures that the changes are minimal and precise by learning when to stop through a special "stopping" action.

In our reinforcement learning model, we construct states based on seeker posts and fixed-length contiguous spans in the associated response posts (Section~\ref{subsec:state}). Insertion, replacement, and deletion of sentences in response posts are defined as actions (Section~\ref{subsec:actions}). We learn a policy that uses transformer language models at its core (Section~\ref{subsec:policy}). We design a reward function that favors empathic, fluent, coherent, specific, and diverse transformations (Section~\ref{subsec:rewards}).

\subsection{State: seeker post \& fixed-length contiguous spans of response post}
\label{subsec:state}
Our agent simultaneously operates on seeker post and fixed-length contiguous spans of response post. The use of seeker post helps us in leveraging conversational context, thereby enabling transformations that are specific to the feelings and experiences shared in the seeker post. The response post is used for making transformations. The use of fixed-length contiguous spans enables a static action set. 

Formally, let $\mathbf{R_{i}}$ contain $n$ sentences $\mathbf{R_{i,1}},...,\mathbf{R_{i,n}}$. At each step, we focus on a contiguous window of $k$ sentences starting from the $j$th sentence $\mathbf{R_{i,j:j+k}} = \mathbf{R_{i,j}},...,\mathbf{R_{i,j+k-1}}$. Then, our state $s \in \mathcal{S}$ is denoted by the pair ($\mathbf{S_{i}}$, $\mathbf{R_{i,j:j+k}}$). Our policy uses a string containing $\mathbf{S_{i}}$ concatenated with $\mathbf{R_{i,j:j+k}}$ separated by a special \textsf{<SPLIT>} token (as commonly used in BERT-like models~\cite{devlin2018bert}).

\subsection{Actions: sentence-level edits}
\label{subsec:actions}

Our agent takes actions at the level of a sentence, i.e. it either inserts new sentences or replaces existing sentences with newer ones. A deletion operation is equivalent to replacing a sentence with an empty string. Our agent can make word-level changes by replacing the original sentence with a slightly different sentence containing only word-level edits. We focus on sentence-level edits because the task of empathic rewriting requires changes that go beyond simple word-level edits. Empathic responses typically contain multiple sentences with different goals such as emotional reactions, interpretations, and explorations~\cite{sharma2020computational}; generating these sentences and using them for making changes to the response is important for empathic rewriting.

In a state ($\mathbf{S_{i}}$, $\mathbf{R_{i,j:j+k}}$), our agent simultaneously takes two actions -- ($a_1$) select a position in $\mathbf{R_{i,j:j+k}}$ for insertion or replacement, ($a_2$) generate a candidate empathic sentence. The action space $\mathcal{A}_1$ of $a_1$ consists of 2k+2 actions -- k+1 positions for insertions, k positions for replacements, and one \textit{special} action for no insertion or replacement, which stops the agent from making any further changes. The action space $\mathcal{A}_2$ of $a_2$ consists of all arbitrary-length sentences. We denote the action taken by our agent as $a$ = ($a_1$, $a_2$) $\in \mathcal{A}_1 \times \mathcal{A}_2$.

\subsection{Policy}
\label{subsec:policy}
At its core, our policy has a transformer language model consisting of a stack of masked multi-head self-attention layers, based on GPT-2 (for a detailed description, see Vaswani et al.~\cite{vaswani2017attention}, Radford et al.~\cite{radford2019language}). It takes as input an encoded representation of our state ($\mathbf{S_{i}}$, $\mathbf{R_{i,j:j+k}}$) and generates the action $a$ = ($a_1$, $a_2$).

\xhdr{($a1$) Selecting a position for insertion or replacement} 
Given ($\mathbf{S_{i}}$, $\mathbf{R_{i,j:j+k}}$) as input, we want to identify a position $\mathbf{p_i}$ in $\mathbf{R_{i,j:j+k}}$ where changes need to be made for improving empathy through insertion or replacement operations. A $k$ sentence window $\mathbf{R_{i,j:j+k}}$ has $k+1$ positions for insertions and $k$ positions for replacement. Then, our task is to select one of these $2k+1$ positions. We formulate this as a classification problem with $2k+2$ classes. The first $2k+1$ classes represent one of the $2k+1$ potential positions and the last class represents the "stopping" action of not selecting any position, thereby stopping the agent from making any changes and keeping the response span unchanged. 

For selecting this position, we first encode the input string "$\mathbf{S_{i}}$ \textsf{<SPLIT>} $\mathbf{R_{i,j:j+k}}$" using the transformer block of GPT-2. We then pass this encoded representation through a linear layer to get the prediction $\mathbf{\hat{p}_i}$ of the position for insertion or replacement. We denote our position classifier as $p_{\text{pos}}$.

\xhdr{($a2$) Generating a candidate sentence} 
Given ($\mathbf{S_{i}}$, $\mathbf{R_{i,j:j+k}}$) as input, we want to generate a candidate sentence $\mathbf{C_{i,j}}$ to be used for making changes to $\mathbf{R_{i,j:j+k}}$. We frame this task as a language modeling problem where the objective is to generate $\mathbf{C_{i,j}}$ that maximizes the conditional probability $p_{\text{sent}}(\mathbf{C_{i,j}} | \mathbf{S_{i}}, \mathbf{R_{i,j:j+k}})$. 

Similar to the position selection action, we first encode our input string "$\mathbf{S_{i}}$ \textsf{<SPLIT>} $\mathbf{R_{i,j:j+k}}$" using the transformer block of GPT-2. We then compute a probability distribution over vocabulary tokens by transforming the encoded representation into a vocabulary-sized vector through a softmax layer. Finally, we use top-p sampling~\cite{holtzman2019curious}\footnote{For generating every word in a sequence, top-p sampling (or nucleus sampling) chooses from the smallest set of words whose total probability is more than p.} over this probability distribution to generate the desired $\mathbf{C_{i,j}}$. The generation is terminated when the sampling process encounters a special end-of-sequence token.

\subsection{Rewards}
\label{subsec:rewards}
Our reward functions aim to increase empathy in posts and maintain text fluency, sentence coherence, context specificity, and diversity:

\xhdr{Change in empathy}
The task of empathic rewriting requires transformations that can increase empathy of posts. Thus, we want to reward actions that increase empathy of $\mathbf{R_{i}}$ and penalize actions that decrease empathy of $\mathbf{R_{i}}$. Let $f_{e}(\cdot)$ be a function that measures empathy of posts. Then, the change in empathy reward, $r_{e}$, is defined as:
\begin{align}
    r_e &= f_e(\mathbf{\hat{R}_{i}}) - f_e(\mathbf{R_{i}})
\end{align}
Here, we estimate $f_{e}(\cdot)$ using the empathy classification model developed by Sharma et al.~\cite{sharma2020computational} for predicting empathy levels of responses. Sharma et al.~\cite{sharma2020computational} leverage a theoretically-grounded framework of empathy consisting of three empathy communication mechanisms (emotional reactions, interpretations, and explorations) and devise a scale of empathy levels from 0 to 6. They train a classification model (RoBERTa~\cite{liu2019roberta}, accuracy $\sim$ 80\%) for predicting empathy of response posts on this scale. We use their trained model as $f_{e}(\cdot)$ which gives us empathy scores of $\mathbf{\hat{R}_{i}}$s in the range of 0 to 6.

\xhdr{Text fluency}
We want to prevent actions that lead to outputs that are highly empathic but not fluent or grammatically correct. Therefore, we want to reward actions that lead to fluent outputs and penalize actions resulting in non-fluent outputs. Here, we operationalize \textit{text fluency} as the inverse of perplexity of the generated $\mathbf{\hat{R}_{i}}$s. We define the text fluency reward, $r_f$ as:
\begin{align}
    r_f &= p_{\text{LM}}\left(\mathbf{\hat{R}_i}\right)^{(1/N)}
\end{align}
where $p_{\text{LM}}$ is a general language model for English and $N$ is the number of words in $\mathbf{\hat{R}_{i}}$. Here, we use GPT-2~\cite{radford2019language} as our $p_{\text{LM}}$, following previous work~\cite{ma2020powertransformer,dai2019style}.

\xhdr{Sentence coherence}
A key component of our action space is the addition of the candidate sentence to the original response. While the candidate sentence might be highly empathic and fluent, it may not be well-suited for the response $\mathbf{R_{i}}$ to which it would be added, leading to incoherent sentences in the transformed response $\mathbf{\hat{R}_{i}}$. This may not be handled by perplexity which tends to give high scores to posts where individual sentences are all fluent but are not coherent at the macro response level. Here, we design a reward function, $r_c$ that measures coherence of the candidate sentence $\mathbf{C_{i,j}}$ with the response span $\mathbf{R_{i,j:j+k}}$. $r_c$ measures the average sentence coherence probability between a candidate sentence and existing sentences in the response. 

First, we create a dataset of likely coherent and incoherent sentence pairs. Given two sentences $\mathbf{R_{i,j1}}$ and $\mathbf{R_{i,j2}}$ in a response $\mathbf{R_i}$, we call ($\mathbf{R_{i,j1}}$, $\mathbf{R_{i,j2}}$) a \textit{potential coherent sentence pair}. We randomly sample a sentence $\mathbf{R}{'}$ which is not a part of responses posted to the current seeker post $\mathbf{S_{i}}$ and call ($r{'}$, $\mathbf{R_{i,j}}$) a \textit{potential incoherent sentence pair} ($\forall \mathbf{R_{i,j}} \in \mathbf{R_i}$). Next, we train a text classification model, based on BERT~\cite{devlin2018bert}, on this dataset. We take softmax at the last layer which gives us probabilities of a sentence pair being coherent ($p_{\text{coherent}}$) or incoherent ($p_{\text{incoherent}}$). Then, our sentence coherence reward is defined as: 
\begin{align}
    r_c &= \frac{\displaystyle\sum_{l=j}^{l=j+k}p_{\text{coherent}}\left(C_{i,j}, \mathbf{R_{i,l}}\right)}{k}
\end{align}
\xhdr{Mutual information for specificity and diversity}
In the process of empathic rewriting, the final rewritten response may become generic (e.g., "\textit{I understand how you feel}") thereby affecting the overall conversation quality~\cite{see2019makes,li2016diversity}. In order to ensure specificity to the seeker post and diversity of responses, we exploit the idea of maximizing mutual information between seeker post and the rewritten response post~\cite{li2016diversity,li2016deep}. Our mutual information reward is:
\begin{align}
    r_{m} &= \lambda_{\text{MI}} * \log \overrightarrow{p}\left(\mathbf{\hat{R}_{i}}|\mathbf{S_{i}}\right) +  (1 - \lambda_{\text{MI}}) * \log \overleftarrow{p}\left(\mathbf{S_{i}}|\mathbf{\hat{R}_{i}}\right)
\end{align}
where $\overrightarrow{p}$ is the transformer language model used in our policy and $\overleftarrow{p}$ is an identical language model for performing the reverse task of generating seeker post from the rewritten response.

\xhdr{Total reward} 
Our total reward is $r = w_{e} * r_{e} + w_{f} * r_{f} + w_{c} * r_{c} + w_{m} * r_{m}$.

\subsection{Optimization and training}
\label{subsec:optimization}

\xhdr{Warm-start using supervised learning}
We use the pre-trained weights of DialoGPT~\cite{zhang2019dialogpt} for initializing our transformer language model. Next, we use a warm-start strategy using supervised learning on a parallel dataset of (low empathy, high empathy) pairs, following previous work in reinforcement learning for dialogue generation~\cite{li2016deep}. For creating this dataset, we follow the reverse process of making highly empathic responses less empathic by removing sentences that are high in empathy. Similar "reverse-engineering" strategy has also been shown to work well in other complex linguistic phenomenon like humor~\cite{west2019reverse}. We first identify highly empathic sentences (with scores $\geq 2$) in our dataset of empathic interactions (Section~\ref{subsec:empathic-data}). For a seeker post $\mathbf{S_{i}}$ and response post $\mathbf{R_{i}}$ having a highly empathy sentence $\mathbf{R_{i,j}}$, we create a dataset with ($\mathbf{S_{i}}$ \textsf{<SPLIT>} $\mathbf{R_{i}}$, $\mathbf{R_{i}} - \mathbf{R_{i,j}}$) pairs.\footnote{$\mathbf{R_{i}} - \mathbf{R_{i,j}}$ refers to the full response post $\mathbf{R_{i}}$ with the sentence $\mathbf{R_{i,j}}$ removed.} We use this dataset to finetune our DialoGPT-initialized transformer language model.

\xhdr{REINFORCE with a baseline value for training}
We use the standard REINFORCE algorithm~\cite{williams1992simple} for training our agent. Our loss function is defined as:
\begin{align}
    J(\theta) &= - (r - b) * \left(\log p_{\text{pos}}\left(a_{1} | \mathbf{S_{i}}, \mathbf{R_{i,j:j+k}}\right) \right. \nonumber \\
    & \left. + \log p_{\text{sent}}\left(a_{2} | \mathbf{S_{i}}, \mathbf{R_{i,j:j+k}}\right) \right) 
\end{align}
where $\theta$ is our set of parameters and $b$ is a baseline estimate of the reward (running average of previous 100 reward values) used for stabilizing training.


\xhdr{Experimental setup} \textb{We use a batch size of 16 and train our model for 20000 steps using a learning rate of 1e-5. We use $w_e = 1.0$, $w_f = 10.0$, $w_c = 0.1$, and $w_m = 0.1$ (selected using a grid-search approach with three values (0.1, 1.0, 10.0) for each hyperparameter). Moreover, we choose $k = 2$, p~$= 0.92$, and $\lambda_{\text{MI}} = 0.5$. We truncate both seeker and response post to 64 tokens each. }
\section{Experiments}
\label{sec:experiments}

\begin{table*}
\centering
\begin{tabular}
{p{1.5cm}p{2cm}|c|c|c|cc|c|c}
\toprule
& \multirow{2}{*}{Model} & \multirow{2}{*}{\parbox{1.7 cm}{\centering Change in empathy ($\uparrow$)}} & \multirow{2}{*}{Perplexity ($\downarrow$)} & \multirow{2}{*}{Specificity ($\uparrow$)} & \multicolumn{2}{c|}{Diversity ($\uparrow$)} & \multirow{2}{*}{\parbox{1.8 cm}{\centering Sentence coherence ($\uparrow$)}} & \multirow{2}{*}{Edit rate ($\downarrow$)} \\
& & & & & distinct-1 & distinct-2 & &  \\
\midrule
\multirow{2}{*}{\parbox{1.5 cm}{\centering Dialogue Generation}} 
& DialoGPT~\cite{zhang2019dialogpt} & 0.4698 & 8.6500 & 0.8921 & 0.0382 & 0.1334 & 0.6683 & 1.3520   \\
& MIME~\cite{majumder2020mime} & 1.2069 & 9.0171 & 0.8837 & 0.0031 & 0.0198 & 0.3687 & 1.8193 \\ 
\midrule
\midrule
\multirow{2}{*}{\parbox{1.5 cm}{\centering Seq-to-Seq Generation}} 
& Latent Seq.~\cite{he2019probabilistic} & 0.9745 & 8.7143 & 0.8512 & 0.0001 & 0.0002 & \textbf{0.9252} & 7.8853  \\
& BART~\cite{lewis2019bart} & -0.0611 & \textbf{7.2040} & 0.8878 & \textbf{0.0722} & \textbf{0.3945} & 0.4560 & \textbf{0.7496} \\
\midrule
\midrule
& \textbf{\ourmodel} & \textbf{1.6410} & 7.3641& \textbf{0.9052} & 0.0659& 0.3807 & 0.3030 & 0.9654 \\
\bottomrule
\end{tabular}
\caption{Performance of \ourmodel and comparisons with dialogue generation and other sequence-to-sequence generation baselines on the set of automatic metrics. \ourmodel outperforms all baselines in empathy improvement and generates fluent, specific, and diverse outputs with lower edits. ($\uparrow$) indicates higher is better, ($\downarrow$) indicates lower is better.} 
\vspace{-15pt}
\label{tab:main-results}
\end{table*}




Next, we present experiments for analyzing the performance of \ourmodel on the task of empathic rewriting. We first describe automatic evaluation metrics (Section~\ref{subsec:automatic-metrics}) based on the desired goals for empathic rewriting (Section~\ref{subsec:goals}), baseline approaches and ablations (Section~\ref{subsec:baselines}), and demonstrate results on the automatic evaluation metrics (Section~\ref{subsec:automatic-results}). Since evaluation using automated metrics in language generation tasks are often not robust~\cite{liu2016not}, we additionally present human evaluation results from people having expertise in therapy and mental health (Section~\ref{subsec:human-eval}). We end with a qualitative discussion on the model's performance (Section~\ref{subsec:qualitative}).

\subsection{Automatic evaluation metrics}
\label{subsec:automatic-metrics}
We use a number of automatic metrics that are based on the goals associated with empathic rewriting (Section~\ref{subsec:goals}):

\begin{itemize}
    \item \textbf{Change in empathy:} A key metric for successful empathic rewriting is how much the empathy has changed from the original response to the rewritten response. Similar to our reward function (Section~\ref{subsec:rewards}), we measure this change using the empathy classification model developed by Sharma et al.~\cite{sharma2020computational}. The model computes empathy scores in the range 0 to 6 (leading to change of empathy ranging from -6 to 6).
    
    \item \textbf{Perplexity:} Similar to our text fluency reward (Section~\ref{subsec:rewards}), we measure perplexity for quantifying fluency of the rewritten responses. For this, we use a pre-trained GPT-2 language model that has not been fine-tuned on our dataset, following previous work~\cite{ma2020powertransformer,dai2019style}. 
    
    \item \textbf{Sentence coherence:} Since empathic rewriting requires changes at the sentence level, ensuring coherent sentences in the final rewritten response is crucial. Here, we measure sentence coherence using the scoring mechanism developed in Section~\ref{subsec:rewards}. 

    \item \textbf{Specificity:} The rewritten response should be specific to the seeker post. Following Xu et al.~\cite{xu2018better}, we measure specificity using word embedding similarity between seeker post and rewritten response post (using embeddings from BERT~\cite{devlin2018bert}).
    
    \item \textbf{Diversity:} Since empathic rewriting has implications on millions of conversations on online mental health platforms, ensuring diversity of responses is important. Here, we measure diversity using the \textit{distinct-1} and \textit{distinct-2} metrics, following Li et al.~\cite{li2016diversity}. The two metrics compute the number of distinct unigrams and bigrams respectively divided by the total number of tokens. 
    
    \item \textbf{Edit rate:} The changes in empathic rewriting should be minimal and precise. Here, we use edit rate~\cite{snover2006study} to measure the number of changes between the original response and the rewritten response. Edit rate is defined by the Levenshtein distance between the two responses divided by the length of the original response.
\end{itemize}

\subsection{Baselines and Ablations}
\label{subsec:baselines}
As the task of empathic rewriting has not been explored before, we compare against baseline approaches from the related tasks of dialogue generation and style transfer. Our baselines are:
\begin{itemize}
    \item \textbf{DialoGPT~\cite{zhang2019dialogpt}:} A large dialogue generation model, based on GPT-2~\cite{radford2019language} and pre-trained on Reddit conversations. 
    \item \textbf{MIME~\cite{majumder2020mime}:} An empathic dialogue generation model which exploits emotion mimicking while accounting for emotion polarity (positive or negative).  
    \item \textbf{Deep latent sequence model~\cite{he2019probabilistic}:} A deep generative model designed for unsupervised style transfer. 
    \item \textbf{BART~\cite{lewis2019bart}:} An encoder-decoder model for sequence-to-sequence language generation. 
    
\end{itemize}

DialoGPT and MIME baselines completely disregard the original response; the rewritten response is the response generated given a seeker post by the respective dialogue generation models. Deep latent sequence model and BART perform a sequence-to-sequence generation from a (seeker post, original response post) pair to a response with higher empathy. We use publicly available implementations of all our baselines. We further fine-tune deep latent sequence model on the dataset of empathy-labeled interactions (Section~\ref{subsec:empathic-data}) and BART on the heuristic-based dataset created for warm-start (Section~\ref{subsec:optimization}). 

Additionally, we investigate the importance of different components of our model using the following ablated baselines:
\begin{itemize}
    \item \textbf{Warm-start only, no RL training:} We analyze the performance of the model at the end of our warm-start stage, i.e. without any RL training.
    \item \textbf{No coherence reward:} We train the model without using the sentence coherence reward.
    \item \textbf{No mutual information}: We train the model without using the mutual information component.
\end{itemize}

\begin{figure}
\centering
\vspace{-10pt}
\subfloat[\ourmodel and MIME are effective at increasing empathy in zero-empathy responses. However, \ourmodel is more effective in increasing empathy in low, non-zero empathic responses and doesn't make an already empathic post worse.]{
	\label{subfig:change-in-empathy-1}
	\includegraphics[width=0.45\columnwidth]{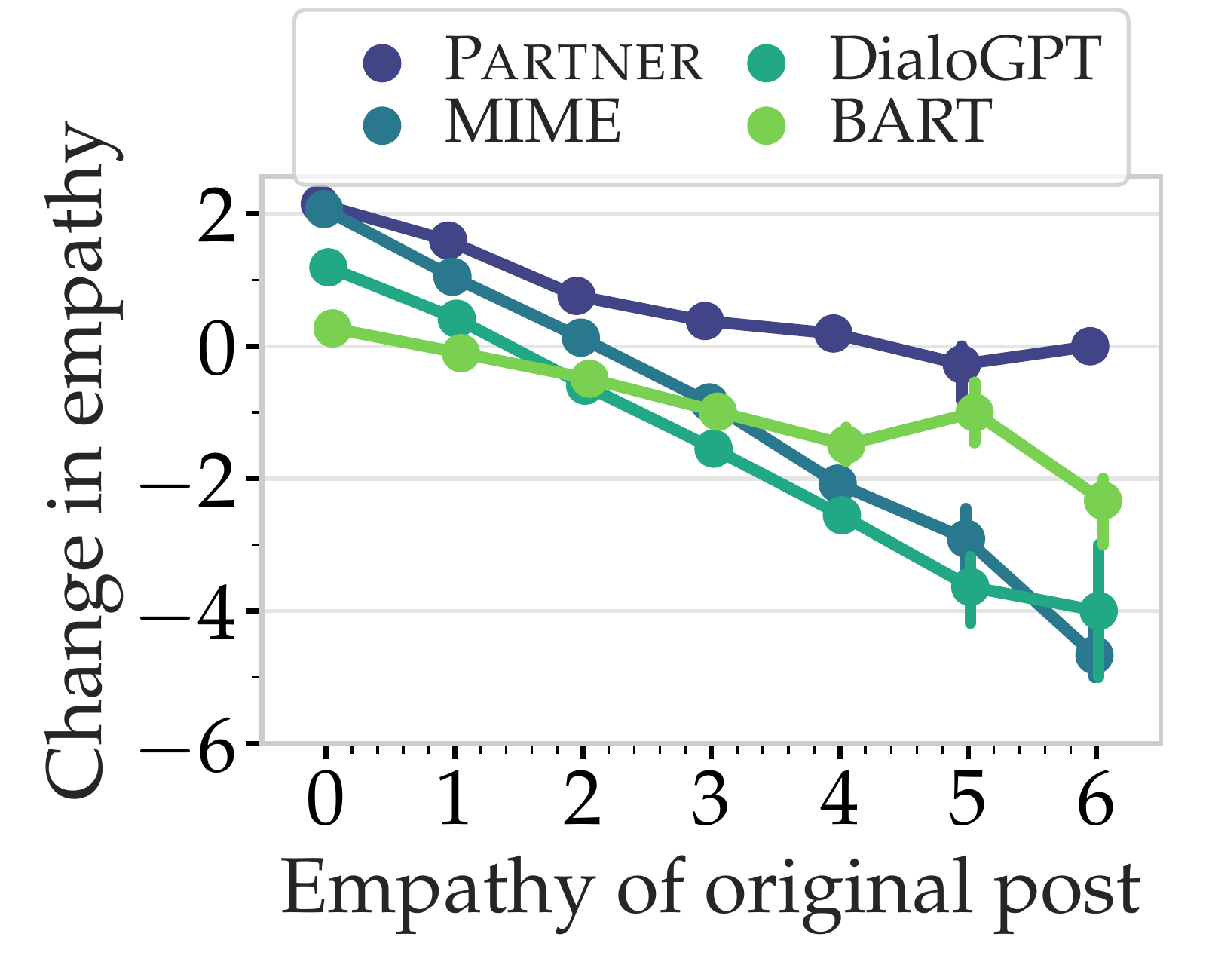} } 
\hfill
\subfloat[\ourmodel makes lesser number of changes compared to baselines. The changes are relatively more for less empathic responses which also tend to be shorter.]{
	\label{subfig:number-of-edits}
	\includegraphics[width=0.45\columnwidth]{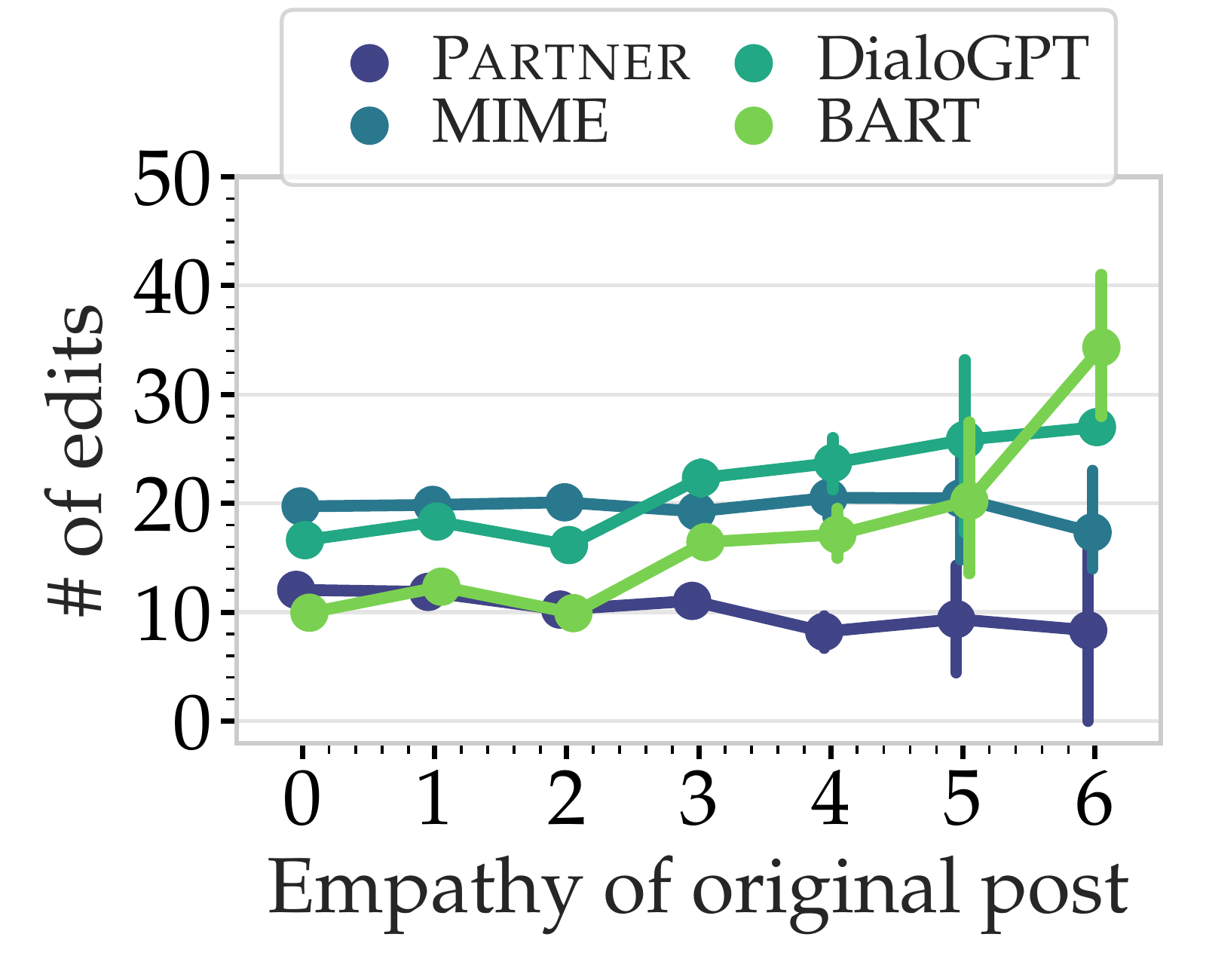} } 
 
\vspace{-10pt} 
\caption{Analysis of empathic rewritings. All error bars in this paper are 95\% confidence intervals.}
\label{fig:rewriting-analysis}
\vspace{-10pt}
\end{figure}

\subsection{Automatic metrics results}
\label{subsec:automatic-results}

\xhdr{Baseline Results} 
Table~\ref{tab:main-results} reports the results of \ourmodel on the automatic evaluation metrics and comparisons with baselines. We find that empathic rewriting through \ourmodel achieves the largest change in empathy (35\% more than the next best approach, MIME) and is more specific than all baselines. MIME generates empathic outputs (+1.21 change in empathy) but the generations have low diversity (86\% less than \textsc{Partner}) indicating similar responses for most seeker posts. BART generates outputs with lowest perplexity, highest diversity, and lowest edit rate, which is consistent with substantial improvements to language models in recent years~\cite{brown2020language}. However, to our surprise, the rewritten responses through BART receive an overall drop of 0.06 in empathy, indicating that the model is unable to perform the task of empathic rewriting well and only generates non-empathic, fluent, diverse text.

Our specificity metric can be hard to interpret with values having a really small range (0.85 to 0.9). However, with human-based evaluation (Section~\ref{subsec:human-eval}), we find that a difference of 0.05 on this metric (between \ourmodel and latent seq.) translates to a 90\% preference towards \textsc{Partner}. \textb{Moreover, while \ourmodel has the lowest sentence coherence score, we find that this is likely due to higher number of sentences generated by it compared to baselines. The baselines generate 1-2 sentence responses on an average, where achieving high coherence between sentences is expected (e.g., a one-sentence response by design has a coherence of 1.0). \textsc{Partner}, on the contrary, generates responses with $\sim$70\% more sentences than baselines, affecting the overall coherence score.} 

\xhdr{Adaptability of rewritings to original post} 
Adapting to different types of original responses and making appropriate changes is an important aspect of empathic rewriting. A low empathic response needs a lot more improvements and edits than a highly empathic response. Figure~\ref{subfig:change-in-empathy-1} shows the change in empathy of responses given their original empathy levels. We find that \ourmodel performs better than baselines in improving responses with low empathy. Importantly, only \ourmodel succeeds at not deteriorating responses that are already highly empathic, indicating the effectiveness of \ourmodel at adapting to responses with different empathy levels. We also analyze the number of edits by each model on responses with different original empathy levels (Figure~\ref{subfig:number-of-edits}). \ourmodel not only effects a greater change in empathy than baselines, it achieves so with the least number of edits for both low and high empathy responses.

\xhdr{Ablation Results}
Table~\ref{tab:ablations} reports results on ablated versions of \textsc{Partner}. Only using warm-start and no RL training is +0.2783 points better than the related off-the-shelf DialoGPT baseline on empathy improvement. However, the RL training in \ourmodel further improves over this warm-start model by +0.8929 points. Using the coherence and mutual information rewards leads to small performance improvements, particularly in empathy (+0.03).

\begin{table*}
\centering
\begin{tabular}
{p{2.2cm}|c|c|c|cc|c|c}
\toprule
\multirow{2}{*}{Model} & \multirow{2}{*}{\parbox{1.7 cm}{\centering Change in empathy ($\uparrow$)}} & \multirow{2}{*}{Perplexity ($\downarrow$)} & \multirow{2}{*}{Specificity ($\uparrow$)} & \multicolumn{2}{c|}{Diversity ($\uparrow$)} & \multirow{2}{*}{\parbox{1.8 cm}{\centering Sentence coherence ($\uparrow$)}} & \multirow{2}{*}{Edit rate ($\downarrow$)} \\
& & & & distinct-1 & distinct-2 & &  \\
\midrule
\textbf{\ourmodel} & \textbf{1.6410} & 7.3641 & 0.9052 & 0.0659 & 0.3807 & 0.3030 & \textbf{0.9654} \\
- no coherence & 1.6127 & 7.2806 & \textbf{0.9055}  & 0.0663 & 0.3844 & 0.3005 & 1.0108 \\
- no mutual info. & 1.6132 
& 7.3274 & 0.9045 & 0.0674 & 0.3859 & \textbf{0.3078} & 1.0071 \\
- warm-start only  & 0.7481 & \textbf{7.1858} & 0.9027 & \textbf{0.0816} & \textbf{0.4238} & 0.2935 & 1.0327 \\
\bottomrule
\end{tabular}
\caption{Ablation results. Warm-start improves over DialoGPT but is still much worse than \ourmodel in empathy improvement, highlighting the effectiveness of our RL-based training.} 
\vspace{-10pt}
\label{tab:ablations}
\end{table*}

\subsection{Human evaluation results}
\label{subsec:human-eval}

Since automatic evaluation in language generation is often not robust~\cite{liu2016not}, we perform a human evaluation on our key metrics (empathy, fluency, and specificity) through A/B testing. We recruit six graduate students in clinical psychology with expertise in empathy and mental health support\footnote{Most participants were PhD students in second or subsequent years of their degree program. Research in Psychology has shown that clinical psychology graduate students are, in general, representative of mental health professionals~\cite{ost2012effects}. Although there are likely some differences between students and licensed psychologists, clinical outcomes in empathy-related measures such as therapeutic alliance have been shown to be comparable while students receive supervision~\cite{goldstein2020outcomes}.} and ask them to compare outputs from \ourmodel against other baseline models, ablations, and expert empathic rewritings (Section~\ref{subsec:empathic-data}) given the same input. Presenting a seeker post, a rewritten response post from \textsc{Partner}, and a rewritten response post from a baseline/ablation/expert-rewrite, we ask them to choose (a) response post which is more empathic, (b) response post which is more fluent, and (c) response post which is more specific. For each model, we collect evaluations on 50-100 examples.

\begin{figure}[t]
\centering
\vspace{-15pt}
\includegraphics[width=\columnwidth]{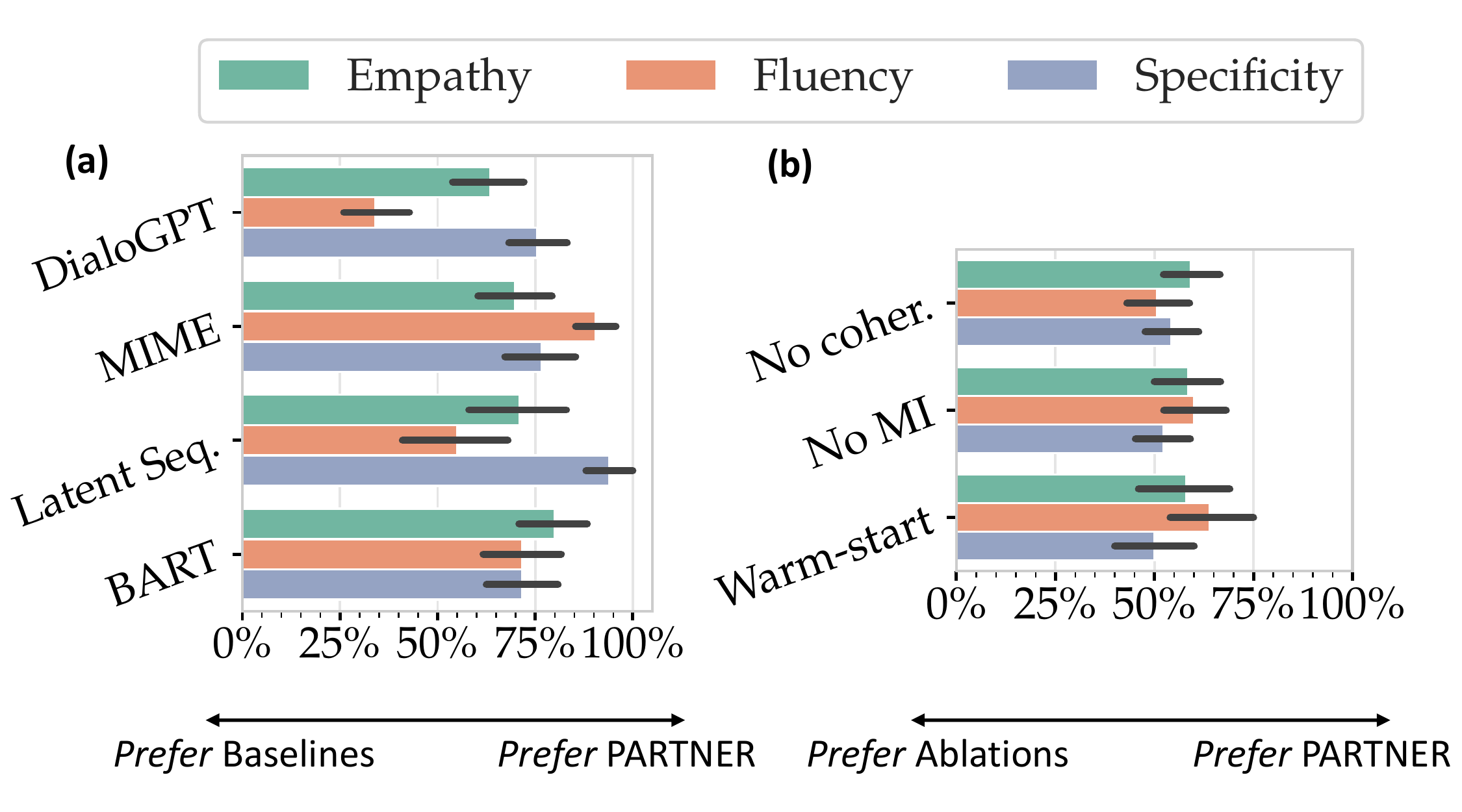}
\vspace{-20pt}
\caption{Human evaluation of empathy, fluency, and specificity in rewritings from \ourmodel vs. (a) rewritings from baseline models, and (b) rewritings from ablations. \ourmodel is preferred over baselines and ablations in empathy and specificity and is competitive in fluency.}
\label{fig:human-eval}
\end{figure}

\xhdr{Results: Baselines and ablations} Figure~\ref{fig:human-eval} shows the percentage of instances in which \ourmodel was preferred over other baselines and ablations (values $>50\%$ indicate preference towards \textsc{Partner}). We find that rewritten responses from \ourmodel are preferred for empathic and specific responses over all baselines. DialoGPT is judged more fluent (Figure 4a) but generates responses following similar templates (e.g., "\textit{I'm sorry you.... I hope you....}"). Moreover, \ourmodel has $\sim$55\% preference for empathy over ablations where coherence and mutual information rewards are not used ($p<0.01$). 

\xhdr{Results: Expert rewritings} 
The most appropriate way of performing empathic rewriting is through human experts. However, experts with training in therapy and mental health support are limited~\cite{olfson2016building} which makes it infeasible to employ them for millions of conversations on online support platforms. We use the small dataset of 180 empathic rewritings from experts to establish what the gold-standard performance for empathic rewritings in mental health support looks like. Unsurprisingly, experts are preferred $\sim$80-90\% times over \ourmodel in empathy, fluency, and specificity ($p<0.001$). However, in 10-20\% cases \ourmodel rewritings are preferred; these are typically instances where \ourmodel is able to make empathic changes to responses while the experts leave it unchanged.

\xhdr{Results: BLEU scores} 
We also use the dataset of expert empathic rewritings (Section~\ref{subsec:empathic-data}) as a ground truth of empathic rewritings and compare outputs of \textsc{Partner}, baselines, and ablations based on this ground truth using the BLEU metric~\cite{papineni2002bleu} (Table~\ref{tab:bleu}). We find that the outputs from \ourmodel are closest to expert rewritings (86\% better than the next best baseline, BART).

\begin{table}[]
\centering
\begin{tabular}{ll|c}
\toprule
 & Model & BLEU score \\ 
 \midrule
 & \textbf{\ourmodel} & \textbf{0.1391} \\ \midrule
 \midrule
\multirow{4}{*}{Baselines} & DialoGPT & 0.0722 \\
 & MIME & 0.0808 \\
 & Latent Seq & 0.0254 \\
 & BART & 0.0956 \\
 \midrule
 \midrule
\multirow{3}{*}{Ablations} & - no coherence & 0.1335 \\
 & - no mutual info. & 0.1297 \\
 & - warm-start only & 0.1281 \\ \bottomrule
\end{tabular}%
\caption{BLEU scores of all models against empathic rewritings from experts. The rewritings from \ourmodel are closest to expert rewritings.}
\vspace{-15pt}
\label{tab:bleu}
\end{table}

\subsection{Qualitative examples}
\label{subsec:qualitative}
We present example rewritings from \ourmodel and baselines in Table~\ref{tab:qualitative-examples}. \ourmodel generates rewritings that leverage both seeker post and original response post for empathic responses. For example, from the seeker post "\textit{I feel like nobody cares about my existence}", \ourmodel is able to infer "\textit{It's hard to find others who can relate}". Also, \ourmodel can coherently transform the response post "\textit{What happened between you two?}" to "\textit{What happened between you two? What caused you to break?}".

\begin{table*}
\centering
\begin{tabular}
{p{3.5cm}p{3.5cm}|p{2cm}|p{6cm}}
\toprule
Seeker post & Response post & Model & Rewritten response post \\ 
\midrule
\multirow{5}{*}[-2em]{\parbox{3.4cm}{I feel like nobody cares about my existence.}} & \multirow{5}{*}[-2em]{I feel the same.} & \ourmodel & It's hard to find others who can relate. I feel the same. \\
\cmidrule{3-4}
& & DialoGPT & I care. \\
\cmidrule{3-4}
& & MIME & That is awful, i am sorry to hear that. I hope you feel better. \\
\cmidrule{3-4}
& & Latent Seq. &  I'm sorry to hear that. \\
\cmidrule{3-4}
& & BART & I think i feel the same way. I do. \\
\midrule
\midrule
\multirow{5}{*}[-2em]{\parbox{3.4cm}{I overthink everything my friends say. I disregard them when they praise me but if they say something negative I take it so personally. I'm sick of being so paranoid.}} & \multirow{5}{*}[-2em]{\parbox{3.4cm}{I understand completely. I feel the same way.}}  & \ourmodel & It's hard when you're with such people and they are being like that to you. I understand completely. I feel the same way. \\
\cmidrule{3-4}
& & DialoGPT & I'm the same way. I'm not sure why, but I just don't like being told I'm being paranoid. \\
\cmidrule{3-4}
& & MIME & That is a shame, i am sure it will not be a bit of a bit! \\
\cmidrule{3-4}
& & Latent Seq. & I know that, i feel the same way i'm sorry to. \\
\cmidrule{3-4}
& & BART & I'm sorry, I understand completely. I feel the same way. I'm here for you. \\
\midrule
\midrule
\multirow{5}{*}[-2em]{\parbox{3.4cm}{I hope I didn't lose my only two best friends. They are the only ones I could relate to.}} & \multirow{5}{*}[-2em]{\parbox{3.4cm}{What happened between you two?}} & \ourmodel & That's awful. What happened between you two? What caused you to break? \\
\cmidrule{3-4}
& & DialoGPT & I hope you didn't lose them. I don't want to lose them. \\
\cmidrule{3-4}
& & MIME & That is a good thing to do. \\
\cmidrule{3-4}
& & Latent Seq. & I'm sorry to hear that. \\
\cmidrule{3-4}
& & BART & What happened? What happened between you two? \\
\bottomrule
\end{tabular}
\caption{Qualitative examples of empathic rewriting using \ourmodel and baseline methods. \ourmodel generates rewritings that leverage both seeker post and original response post for empathic responses.  It infers perspective from seeker posts ("\textit{I feel like nobody cares about my existence}" $\rightarrow$ "\textit{It's hard to find others who can relate}") and generates empathic sentences which can be coherently inserted to response posts ("\textit{What happened between you two?}" $\rightarrow$ "\textit{What happened between you two? What caused you to break?}")} 
\vspace{-10pt}
\label{tab:qualitative-examples}
\end{table*}
\section{Discussion and Conclusion}
The burden of mental illness globally is overwhelming, and common mental disorders are some of the most debilitating illnesses worldwide~\cite{collins2011grand}. Existing mental health resources and interventions are ill-suited to the size of the need. Online mental health support platforms that make use of peer supporters is one route to scaling up support, but the biggest challenge is to effectively train or scaffold the peer supporters. Our empathic rewriting approach represents a foundational proof-of-concept of how computational methods may help peer supporters online.

Rewriting human-generated responses may be an effective approach to balancing the benefits and risks of using artificial intelligence in mental health settings. By combining human knowledge of context and experience, our approach can both provide feedback to online peer-supporters with actionable, real-time examples, and provide support seekers with more empathic responses. Importantly, this machine-in-the-loop approach can help mitigate some of the risks related to toxicity and safety of AI systems in settings of suicidal ideation, self-harm, or insensitive comments related to race/ethnicity/gender~\cite{li2020developing,luxton2012social,collings2012suicide}.

\xhdr{Summary of contributions}
Our work proposes a new task of empathic rewriting for transforming low-empathy conversational posts in online mental health support platforms to higher empathy. For this task, we develop and train \textsc{Partner}, a reinforcement learning model which makes sentence-level edits to posts for making them empathic. Through extensive experiments based on automatic and human evaluation, we show that \ourmodel can effectively generate more empathic posts and outperforms baseline methods from related tasks.

\section*{Acknowledgments}
\textb{We would like to thank TalkLife and Jamie Druitt for their support and for providing us access to a TalkLife dataset. We also thank the members of UW Behavioral Data Science Group and the anonymous reviewers for their suggestions and feedback. This research has been supported in part by a Microsoft AI for Accessibility grant, the Allen Institute for Artificial Intelligence, NSF grant IIS-1901386, and Bill \& Melinda Gates Foundation (INV-004841). A.S.M. was supported by grants from the National Institutes of Health, National Center for Advancing Translational Science, Clinical and Translational Science Award (KL2TR001083 and UL1TR001085) and the Stanford Human-Centered AI Institute. D.C.A. was supported in part by an NIAAA K award (K02 AA023814).}

\xhdr{Conflict of Interest Disclosure} \textb{D.C.A. is a co-founder with equity stake in a technology company, Lyssn.io, focused on tools to support training, supervision, and quality assurance of psychotherapy and counseling.}

\bibliographystyle{ACM-Reference-Format}
\bibliography{0_ref}

\end{document}